\documentclass[runningheads,a4paper]{llncs}

\usepackage{times}
\usepackage{amssymb}
\usepackage{graphicx}
\usepackage{url}
\usepackage{makecell}
\usepackage{multirow}
\usepackage{hyperref}
\usepackage{cleveref}

\newcommand{\keywords}[1]{\par\addvspace\baselineskip
\noindent\keywordname\enspace\ignorespaces#1}

\urldef{\mailsa}\path|{anastasia.zhukova,gipp}@uni-goettingen.de|
\urldef{\mailsb}\path|christian.matt@eschbach.com|

\begin{document}

\title{Efficient Domain-adaptive Continual Pretraining for the Process Industry in the German Language}

\titlerunning{Efficient Domain-adaptive Continual Pretraining}

\author{Anastasia Zhukova\inst{1} \and Christian E. Matt \inst{2} \and Bela Gipp\inst{1}}

\institute{University of G{\"o}ttingen \\
\url{https://gipplab.org/} \\
\mailsa\\
\and
eschbach GmbH \\
\url{https://www.eschbach.com/} \\
\mailsb\\
}

\index{Zhukova, Anastasia}
\index{Matt, Christian E.}
\index{Gipp, Bela}

\toctitle{} \tocauthor{}

\maketitle

\begin{abstract}
Domain-adaptive continual pretraining (DAPT) is a state-of-the-art technique that further trains a language model (LM) on its pretraining task, e.g., masked language modeling (MLM), when common domain adaptation via LM fine-tuning is not possible due to a lack of labeled task data. Although popular, MLM requires a significant corpus of domain-related data, which is difficult to obtain for specific domains in languages other than English, such as the process industry in the German language. This paper introduces an efficient approach called ICL-augmented pretraining or ICL-APT that leverages in-context learning (ICL) and k-nearest neighbors (kNN) to augment target data with domain-related and in-domain texts, significantly reducing GPU time while maintaining strong model performance. Our results show that the best configuration of ICL-APT performed better than the state-of-the-art DAPT by 28.7\% (7.87 points) and requires almost 4 times less GPU-computing time, providing a cost-effective solution for industries with limited computational capacity. The findings highlight the broader applicability of this framework to other low-resource industries, making NLP-based solutions more accessible and feasible in production environments.
\end{abstract}

\section{Introduction}
In Natural Language Processing (NLP), a low-resource language refers to one that lacks sufficient linguistic data, resources, or tools for training and developing NLP models \cite{hedderich-etal-2021-survey,chu-wang-2018-survey}. 
Domain-specific German, particularly in fields such as the process industry that are rich in professional jargon, codes, acronyms, and numerical data, falls into this category due to the limited availability of large, publicly accessible datasets \cite{Zhukova2024}. 
Consequently, language models explicitly tailored to these specialized areas are scarce. 

The process industry refers to sectors involved in continuously producing goods through chemical, biological, or physical processes. This includes industries like chemicals and pharmaceuticals, where publicly available data in these domains in the German language is considerably scarcer than in English, because most of the publications in these areas are in English, and the proprietary data is limited to the collected volumes.

\begin{figure}
    \centering
    \includegraphics[width=0.7\linewidth]{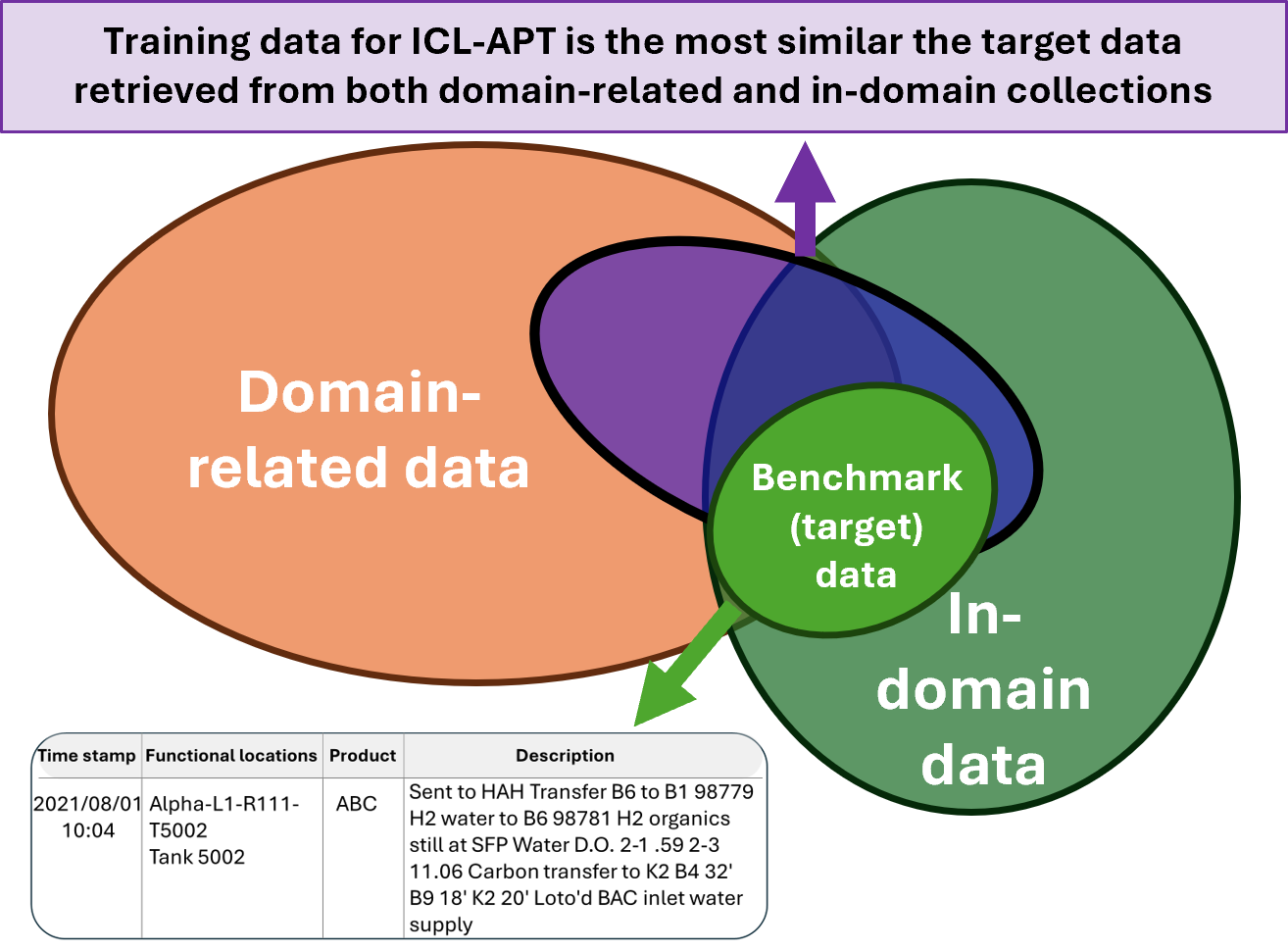}
    \caption{The proposed ICL-APT methodology retrieves data for continual pretraining using target data (light green) from the domain-related document collection (orange) and in-domain data (dark green) to create augmented target data for pretraining. 
    The target data consists of the text logs from the German process industry, which contain domain-specific terminology that demands specialized knowledge of the field and a thorough understanding of the production process.
    }
    \label{fig:diff_data}
\end{figure}

Domain-adaptive continual pretraining (DAPT) is a state-of-the-art technique for further training a language model (LM) on its original task, such as masked language modeling (MLM)~\cite{GururanganMSL20}. While fine-tuning an LM for a specific downstream task is the most common approach for LM domain adaptation, it falls short when labeled data for a domain-specific task is limited or unavailable. 
Unlike model fine-tuning, DAPT does not modify the model architecture by building additional layers on top of an LM; instead, it continues the pretraining process by using domain-specific data. 
According to \cite{GururanganMSL20}, an LM can learn in-domain semantics by training on domain-related data after pretraining on general-domain text. 
As a variation of DAPT, \cite{GururanganMSL20} achieved optimal performance on downstream tasks by further training an LM using human-curated task data, specifically in-domain text that matches the target data distribution. 
Their findings reveal that pretraining on a smaller subset of domain-specific data can significantly improve performance compared to relying on large corpora of broadly domain-related data.

While DAPT remains a state-of-the-art approach, it typically demands large amounts of domain-specific text, often gigabytes of topic-related data, and substantial computational resources like GPUs. Low-resource languages, to which belong high-resource languages within narrow domains such as the process industry in German, cannot reliably use DAPT. Instead, they require methods that work effectively with limited resources while still delivering comparable model performance~\cite{joshi-etal-2020-state}. 

To tackle the challenges of pretraining and fine-tuning models in low-resource settings, prior work has explored optimizing the training process under budget constraints \cite{BaiRX21}. This involves balancing the size of unlabeled data used for continual pretraining, which affects GPU usage, and the size of labeled data for downstream tasks, which influences human annotation costs. However, relatively little research has focused on how to obtain pretraining data that closely resembles the target task data, ensuring a similar distribution between pretraining and fine-tuning datasets~\cite{GururanganMSL20,XieSML23}. 

This paper introduces in-context learning adaptive pretraining \textit{ICL-APT}, an efficient domain-adaptive continual pretraining method that \textit {explores a set of steps for collecting, selecting, and utilizing domain-related (DR) and in-domain (ID) data to augment target datasets}. Specifically, we explore how providing additional context for domain-specific vocabulary can enable cost-efficient adaptive pretraining of LMs using smaller but more domain-rich training text and fewer GPU resources and answers a \textit{\textbf{RQ}: How do data augmentation techniques for domain adaptation via continual pretraining influence the performance of the semantic search task in a specific domain?}. Our experiments showed that the best configuration of ICL-APT outperformed DAPT on the semantic search task by 28.7\% (7.87 points) while requiring almost 1/4 of the GPU hours. We conclude that augmenting domain-specific terminology and vocabulary provides sufficient context for ICL with DR and ID data \cite{GunasekarZAM23,CaciularuCBP21}.

\section{Background}
\label{sec:related_work}

The proposed methodology of ICL-APT lies at the intersection of the two methodologies: continual pretraining and in-context learning. Both continual pretraining and in-context learning rely on the available unlabeled data to facilitate LM domain-adaptation and then rely on transfer learning when fine-tuned on a task-specific dataset. Below, we explain the details of these methodologies to lay the foundation for our methodology.

\textbf{Continual pretraining} \hspace{0.3cm}
Gururangan et al. \cite{GururanganMSL20} introduced a state-of-the-art methodology for continual pretraining of language models (LMs) by training them on large domain-related corpora, enabling domain shifts across various fields, for example biomedical texts in Spanish \cite{ChizhikovaLCM23,RojasDV22} or chemistry in English \cite{JiangJXZ23,KeSLK23}. Gururangan et al. compared various pretraining strategies, emphasizing that data distribution, rather than data volume, is key to effective domain adaptation. They explored domain-adaptive pretraining (DAPT) using domain-related data, task-adaptive pretraining (TAPT) using task-specific data, and human-curated task-adaptive pretraining (cTAPT) using in-domain data. Their findings showed that while DAPT used more training data, it performed similarly to cTAPT in some tasks and domains, but even underperformed in others. This demonstrated that the alignment of data distribution with the task at hand can be more important than the sheer volume of data. They proposed augmented TAPT (aTAPT) for situations with limited human-curated data, which uses k-nearest neighbors (kNN) to retrieve relevant training data from domain-related corpora, achieving comparable results to DAPT while requiring fewer resources.

\textbf{In-context learning} \hspace{0.3cm}
The in-context learning (ICL) is a commonly known technique for the zero- or few-shot learning by prompting LLMs with the specific context and/or examples to follow \cite{brown-2020-language}. ICL is widely applied to shift large language models (LLMs) to specific domains to improve the LLMs' performance on the domain-specific tasks \cite{mosbach-etal-2023-shot}%
. The most recent approaches incorporate retrieval-augmented prompt generation, helping the model learn infrequently seen terms encountered during pretraining and automating domain adaptation of LLMs \cite{LongWP23,MonajatipoorYSE24,GargMS24}. 

Although ICL is a technique applied during inference rather than training, \cite{CaciularuCBP21} demonstrated that pretraining a language model on concatenated documents related to the same narrow topic or event can improve performance on tasks involving document similarity, such as text summarization, semantic textual similarity, and cross-document coreference resolution. This approach enhances the model’s ability to understand terms and phrases within a specific context, producing effects similar to those seen with ICL.

\section{Methodology: ICL-APT}
\label{sec:methodology}

The methodology of \textit{ICL-augmented pretraining (ICL-APT)} revisits ICL to utilize retrieved data that closely resembles the target data, expanding the context for domain-specific terminology for continual pretraining (see \Cref{fig:diff_data}). By incorporating additional context, such as definitions and explanations of abbreviations, LMs develop a deeper understanding of domain semantics during further training tasks like masked language modeling. This augmented data improves the effectiveness and efficiency of the training process. Still, its success relies on the semantic similarity between the training data and the target data, which defines the specific domain.
Our approach builds upon the methodologies of aTAPT \cite{GururanganMSL20} and ICL \cite{LongWP23}, with two primary objectives: (1) transferring most of the data preparation to the CPU, and (2) reducing training time while enhancing the language model's performance on downstream tasks.

\textbf{kNN} \hspace{0.3cm} We enhance the data selection process for domain-related data using kNN, as proposed by \cite{GururanganMSL20}. Our approach keeps kNN lightweight while capturing more context-dependent semantics than the previously used VAMPIRE word embeddings. For implementation, we utilize a sentence-transformers model \textit{paraphrase-multilingual-MiniLM-L12-v2} with the 384 dimension that supports the German language and delivers high performance in document encoding \cite{ReimersG20}. 
We also use a linear search in FAISS with cosine distance on the normalized vectors, but through LangChain integration. 
LangChain provides the flexibility to configure both the number of nearest neighbours and the maximum distance threshold for retrieving neighbours and adds control in more precise retrieval, enhancing the effectiveness of data augmentation.

\textbf{Sources} \hspace{0.3cm} We extend the data selection process with kNN to include both domain-related (DR) and in-domain data (ID) rather than relying solely on DR data (see \Cref{sec:experiments} for details). 
Augmenting target data with ID data is essential because the proprietary data is not present in the publicly available data to enable effective domain adaptation.

\textbf{ICL} \hspace{0.3cm} ICL adds valuable context for new or previously infrequently encountered domain terms in the target data, i.e., a broader context with paragraphs of factual depth. Extended context is essential for learning these terms through masked language modeling (MLM), especially for the tasks that learn additional semantics from the related texts, e.g., cross-document coreference resolution \cite{CaciularuCBP21}. ICL is also utilized in fine-tuning large language models (LLMs) and retrieval-augmented generation for domain-specific tasks. This approach helps improve the quality of results by grounding them in provided instructions and factual information, reducing the risk of hallucinations \cite{LongWP23}. 
Given that text logs in shift books typically range from a sentence to a paragraph in length, ICL can be applied by concatenating the top nearest neighbours while keeping the text within the 512-token input limit of most language models.

\textbf{Masked language modeling setup} \hspace{0.3cm} Following the approach of \cite{GururanganMSL20}, we create several versions of the randomly masked tokens within the same text. By iterating over the text with varied masked tokens, the model is exposed to multiple domain terms within the same context, enhancing its ability to learn them effectively. In ICL-APT, we enhance the context by concatenating the top nearest neighbors to the target data, thereby expanding the context from which the language model can learn.

\textbf{Pipeline} \hspace{0.3cm} A dataset for ICL-APT is constructed as follows. First, we retrieved K NNs for both DR and ID data types with a maximum cosine distance of D. The retrieved NNs of each source were concatenated to the corresponding seed records in increasing order of cosine distance. Hence, for one seed record, we have two context-augmented texts. If no data was retrieved for augmentation, the seed record was used as-is. Second, we randomly mask tokens in each context-augmented text X or Y times, depending on the source, i.e., ID or DR, and all these versions of the masked texts are added to the train set. Different masking ensures that several domain-specific tokens can be learned within the same context.  

\section{Experiments}
\label{sec:experiments}

\subsection{Experiment setup}
\textbf{Model} \hspace{0.3cm} 
We use GBERT-base as a German-only LM with the number of parameters under 150M \cite{ChanSM20}. GBERT tokenizer is designed with a larger vocabulary of German words and can effectively handle long tokens by preserving meaningful components in compound words. 
For instance, if the compound word \textit{Wasserpumpe} (water pump) is not present in the vocabulary, it is split into two tokens: \textit{Wasser} and \textit{\#\#pumpe}. 
We exclude LLMs from the consideration since running LLMs for inference requires GPUs, and BERT-like models can run sufficiently on CPUs. 
 
\textbf{Target data} \hspace{0.3cm} We define the process industry domain using text-only data from a custom domain benchmark designed for downstream classification tasks based on shift logs. These text logs are sourced from multiple shift books to ensure diversity and have been manually verified for quality. Unlike \cite{GururanganMSL20}, who used the term "task data" to describe domain adaptation for a single NLP task, we use the term \textit{target data} to refer to text data spanning multiple NLP tasks within the domain benchmark. This term reflects our distinctive approach, where multiple tasks collectively define the domain. The target data serves as a reference point for identifying similar tasks to facilitate data augmentation. 
We utilized a sample of 10K text logs for the experiments. 

\textbf{Source data} \hspace{0.3cm} In our experiments, we used two types of source data: \textit{domain-related} (DR) and \textit{in-domain} (ID). We define DR data for the process industry as the overlap between the chemistry, pharmaceuticals, and engineering domains. We collected 10,3 GB of German domain-related data from publicly available sources, such as dissertations \cite{dnb_diss}\footnote{\url{https://github.com/anastasia-zhukova/dissertations_parser_DNB}}, patents \cite{ddr_patents,eu_patents}, Wikipedia, and EU regulations \cite{regulations_chemistry,regulations_pharma,regulations_machinery}. ID data consists of 3.2M text logs (1.51GB) from twenty shift books in the chemical and pharmaceutical domains. Each \textit{text log} is a short paragraph-sized text that describes events, statuses, directives, or tasks in a production plant.

\textbf{Semantic search task} \hspace{0.3cm} We evaluate ICL-APT on a semantic search task using zero-shot learning, following the approach of the BEIR benchmark \cite{beir_2021}. First, the continually-pretrained model, its modifications, and baselines are evaluated as-is without any fine-tuning as a text encoder. Second, we use the BEIR framework to fine-tune the continually pre-trained model as a text encoder. To train a bi-encoder, we use a DR version of German \href{https://huggingface.co/datasets/microsoft/ms_marco}{MS MARCO} \cite{Tri2016}, which is created with a binary classifier. \footnote{
This classifier is a fine-tuned SciBERT model \cite{BeltagyLC19} trained on an 80K dataset that combines \href{https://huggingface.co/datasets/allenai/scirepeval/viewer/fos}{fields of study (FoS)} data and our ID data. The train dataset was split so that 50\% consisted of DR texts, selected by the labels that correspond to our domain, and ID texts, thereby forming the true labels, while the other 50\% was sampled from various areas of FoS.}
The resulting dataset contains approximately 2.27M training pairs (132K positive).

We use an ID test collection based on seven shift books, resulting in 205 search queries over a collection of 330K documents \cite{zhukova-etal-2025-automated}. Some portions of the text data from this collection may appear in both the target and source datasets. This setup adheres to the definition of TAPT, where text-only task data is used for continual pretraining, followed by task-specific fine-tuning using labeled data. 

\textbf{Metrics} \hspace{0.3cm} To evaluate semantic search's retrieval and ranking capabilities, we compute a mean of three metrics for information retrieval: mean average precision (MAP@10), mean reciprocal rank (MRR), and normalized commutative normalized discounted cumulative gain (nDCG@10). We report results as a mean across the seven shift books. To provide a comprehensive comparison across all setups, we report both the number of training steps and the required GPU hours. 

\textbf{Baselines} \hspace{0.3cm} First, we compare ICL-APT to GBERT-base and the original configurations of continual pretraining proposed by \cite{GururanganMSL20}, which include: (1) DAPT: Pretraining on the full domain-related (DR) data, (2) cTAPT: human-curated task-adaptive pretraining on the full ID data, (3) TAPT: the text-only data from the labeled benchmark data, (4) DAPT + cTAPT: cTAPT applied after pretraining on DAPT, (5) DAPT + TAPT: TAPT applied after pretraining on DAPT. 

Second, the baselines are publicly-available bi-encoders under 150M parameters: \href{https://huggingface.co/sentence-transformers/paraphrase-multilingual-MiniLM-L12-v2}{sentence-transformers/paraphrase-multilingual-MiniLM-L12-v2} (SentTr-1), \href{https://huggingface.co/sentence-transformers/msmarco-distilbert-multilingual-en-de-v2-tmp-lng-aligned}{sentence-transformers/msmarco-distilbert-multilingual-en-de-v2-tmp-lng-aligned} (SentTr-2), and \href{https://huggingface.co/ibm-granite/granite-embedding-107m-multilingual}{ibm-granite/granite-embedding-107m-multilingual} (IBM-Granite). Additionally, we compare the results to the state-of-the-art \href{https://huggingface.co/intfloat/multilingual-e5-large}{multilingual E5-large} (mE5) text encoder (560M parameters). 

\textbf{Implementation details} \hspace{0.3cm} We used one A100 v4 (24 vCPUs, 220 GiB memory) with fixed training parameters, which are reused from \cite{GururanganMSL20}: probability of MLM of 0.15, batch size of 64 (the batch size is modified to fit a single A100), gradient accumulation steps of 1,  Adam epsilon of 1e-6, weight decay of 1e-2, and warm-up ratio of 6e-2. We adopt an implementation from HuggingFace to perform continual pretraining with MLM. We filled out the full capacity of 512 input tokens and truncated extra or padded missing tokens.  

We trained DAPT for one epoch in line with the original experiment of \cite{GururanganMSL20}. However, since our cTAPT corpus is significantly larger than the original setup, we also trained cTAPT for just one epoch. While the original configuration for aTAPT involved 100 epochs, we trained ICL-APT for 20 epochs to align with a low-computation-power scenario. Similarly, TAPT was trained for 20 epochs. To evaluate the influence of the number of epochs, we additionally trained TAPT for 80 epochs, which resulted in a comparable training time to ICL-APT. In our experiments, we use K=3 for kNN but evaluate several configurations of data for ICL-APT. For example, a configuration "10:10, 0.8" means that text records for the context-augmentation were retrieved with a cosine distance of 0.8, both ID-augmented and DR-augmented text have 10 masking variations, and it results in a dataset of 200K documents. To evaluate a text encoder, we use a SentenceTransformers wrapper \cite{ReimersG20} for BERT models for document encoding, which applies mean pooling to the output vectors. All vectors are normalized during indexing and retrieval to ensure consistent performance.

\subsection{Results and Discussion}

In the original study, \cite{GururanganMSL20} demonstrated that cTAPT and combinations such as DAPT+ cTAPT or DAPT+TAPT achieved the best performance across multiple tasks from various domains. Our experiments investigate whether the same results can be achieved in the process industry domain. \Cref{tab:results_baselines} shows that the proposed ICL-APT achieved the best overall performance of 35.28 and outperformed the state-of-the-art setting of DAPT by 28.7\%. Among baselines, TAPT trained for 80 epochs performed the second best (33.86), and DAPT+TAPT performed the third best, both given even longer training time compared to the best performing configuration of ICL-APT. We see that our results do not show any superiority of cTAPT, DAPT+cTAPT, or DAPT+TAPT, given the described experiment setup for the low-resource model training and rather limited data availability of the ID or DR data of production logs. We hypothesize that the language used in these logs is highly specialized, containing numerous codes, jargon, abbreviations, and incomplete syntax. As a result, further training without providing additional context for this specific terminology fails to deliver the expected performance.

\begin{table*}
\centering
\small
\begin{tabular}{l|l|r|l|r|r|r|r|r} 
\hline
Model & \makecell{Parameters\\ (DR : ID,\\ cos. dist.) }  & \makecell{Train data\\ size (GB)}  & \makecell{Total steps/\\ Epochs} & GPU &  MAP@10 & MRR & nDCG@10 & Mean \\
\hline
GBERT-base & -- & -- & -- & -- &  21.42 & 24.12 & 9.18  & 18.24 \\
\hline
DAPT & -- & 10.30 &233K / 1 & 22h &  31.89 & 34.52  & 15.83  & 27.41 \\
cTAPT & -- & 1.51 & 41K  / 1 & 4h & 29.45 & 32.29 & 14.77  & 25.50 \\
TAPT & -- & 0.01 & 10K / 20 & 1h &  34.27 & 37.75 & 18.37  & 30.13 \\
TAPT & -- & 0.01 & 147K / 80 & 6h & \textit{\underline{37.18}}  & \textbf{\textit{43.19}} & \textit{\underline{21.22}}  & \textit{\underline{33.86}} \\
DAPT+cTAPT & -- & 11.81 &  274K / 1+1 & 26h &  29.37 & 32.91 & 14.64  & 25.64 \\
DAPT+TAPT  & -- & 10.31 & 243K / 1+20 & 23h &  35.72 & 39.43 & 18.42  & \underline{31.19}\\
\hline
\multirow{5}{*}{ICL-APT} 
        & 10 : 10, 0.8 & 0.12 & 56K / 20 & 6h &  36.37 & 40.29 & 19.43  & 32.02\\
        & 10 : 10, 0.7 & 0.12 & 28K / 10 & 3h & 37.10 & 40.27 & 20.05 & 32.47 \\
        & 10 : 10, 0.7 & 0.12 & 42K / 15 & 4.5h & \textbf{40.32} & \textit{\underline{43.02}} & \textbf{21.80} & \textbf{\textit{35.04}} \\
        & 20 : 10, 0.7 & 0.18 & 42K / 10 & 4.5h & 37.09 & 39.92 & 19.37 & 32.13 \\
        & 10 : 20, 0.7 & 0.18 & 42K / 10 & 4.5h & \textbf{\textit{39.54}} & \textbf{44.73} & \textbf{\textit{21.57}} & \textbf{35.28}\\
\hline
\end{tabular}
\caption{The proposed ICL-APT not only outperforms the baselines from the state-of-the-art methods, e.g., DAPT, TAPT, and cTAPT, but also performs continual pretraining with fewer GPU hours. A stricter threshold for the data selection and applying a larger number of masked tokens variations to in-domain (ID) data yields the best performance across multiple configurations of ICL-APT.}
\label{tab:results_baselines} 
\end{table*}

Moreover, \Cref{tab:results_baselines} shows the positive impact of the following parameters. First, using a stricter cosine distance threshold when retrieving the K nearest neighbors for ICL improves the quality of the context when augmenting seed records, leading to better performance even with twice as few training epochs (32.47 with a threshold of 0.7 vs. 32.02 with a threshold of 0.8). Second, applying a larger number of token masking variations to the ID text compared to DR test results in better performance (35.28 using DR=10 and ID=20 vs. 32.13 using DR=20 and ID=10). Third, as expected, using a larger number of training steps or epochs improves performance when the same training data is used (35.04 for 15 epochs vs. 32.47 for 10 epochs). The results indicate that the best-performing configuration used a small cosine distance, a larger number of token masking variations in the ID data, and fewer epochs compared to other configurations. This suggests that the quality of the training data and the ability to learn new domain-specific terms (i.e., through context and multiple token masking variations) are more important than the number of epochs required for training the model.

\Cref{tab:ablation} reports the results of the ablation studies and shows the positive contribution of each component of ICL-APT. To maintain consistency across setups with reduced components, we limited the maximum training data size to 200K records and ensured that the product of $training\_size \times epochs$ remained between 400K and 640K. This setting guarantees that the largest dataset is trained for at least three epochs. For setups involving DR and ID data without kNN-based data selection, we randomly sampled 200K records from each dataset. We observed that the most significant contribution in learning domain-specific terms comes from the large variations of the token masking, and it confirms the findings obtained from \Cref{tab:results_baselines}.

\begin{table}[htbp]
  \centering
  \begin{minipage}{0.4\textwidth}
  \centering
        \begin{tabular}{l|r|r} 
        \hline
        Models & Mean & $\delta$  \\
        \hline
        ICL-APT (10:10, 0.8) & 32.03 & --\\
        \hspace{2mm} --diff.token.mask. & 27.98  & -4.05 \\
        \hspace{2mm} --20 epochs  & 27.05 & -0.92 \\
        \hspace{2mm} --ICL & 25.60 & -1.46\\
        \hline
        \hspace{2mm} (ID only)    & & \\
        \hspace{7mm} --DR-kNN & 23.87 & -1.73 \\
        \hspace{7mm} --DR & 23.87 & -2.03\\
        \hline
        \hspace{2mm} (DR only) & &\\
        \hspace{7mm} --ID-kNN  & 20.25 & -5.35\\
        \hspace{7mm} --ID & 23.73 & +3.48 \\
        \hline
        \end{tabular}
        \caption{Ablation studies of ICL-APT: each component positively impacts the proposed methodology. }
        \label{tab:ablation}
  \end{minipage}
  \hfill
  \begin{minipage}{0.58\textwidth}
  \centering 
        \begin{tabular}{l|c|r|r|r|r}
        \hline
        Models & dFT & MAP@10 & MRR & nDCG@10 & Mean \\
        \hline
        SentTr-1 & - & 43.84 & 46.99 & 26.90 & 39.24 \\
        SentTr-2& - & 54.83 & 59.81 & 35.12 & 49.92 \\
        mE5 (560M) & - & 59.82 & 65.31 & 42.26 & 55.80 \\
        IBM-Granite & - & \textit{\textbf{62.42}} & \textbf{\textit{67.45}} & \underline{\textit{44.67}} & \underline{\textit{58.18}} \\
        \hline
        GBERT & + & \textbf{62.64} & \underline{\textit{66.92}} & \textbf{\textit{45.78}} & \textbf{\textit{58.45}} \\
        TAPT (80 ep) & + & 61.26 & 65.63 & 42.43 & 56.44 \\
        DAPT+TAPT & + & 60.95 & 66.28 & 43.06 & 56.76 \\
        ICL-APT & + & \underline{\textit{62.21}} & \textbf{69.36} & \textbf{45.81} & \textbf{59.13} \\
        \hline
        \end{tabular}
        \caption{Evaluation of models domain fine-tuned (dFT) as bi-encoders using DR dataset. ICL-APT outperforms the models continually pretrained with other continual pretraining techniques and publicly available text encoders of comparable model size. }
        \label{tab:bi-encoder}

  \end{minipage}
\end{table}

Lastly, we evaluate the capability of ICL-APT  and its baselines for domain transfer learning when fine-tuned as bi-encoders using a DR task-specific dataset. We fine-tuned all models for 5 epochs. \Cref{tab:bi-encoder} shows that ICL-APT fine-tuned on the DR version of MS MARCO outperformed other models after continual pretraining. Although TAPT (80 epochs) and DAPT+TAPT outperformed GBERT-base when evaluated without fine-tuning, their performance did not proportionally grow after fine-tuning, which suggests a decline in the capabilities of transfer learning after continual pretraining. While ICL-APT outperforms GBERT by the mean of metrics of 0.68, it yields the best MRR (i.e., the position of the first hit). Moreover, ICL-APT outperforms the state-of-the-art multilingual E5, despite being five times smaller, and IBM-Granite, while, by estimation, requiring less than 80 times the training data for all training steps.

While our experiments demonstrate significant improvements in the semantic search task using ICL-APT, the availability of high-quality in-domain and domain-related data plays a crucial role in optimizing performance across different domains and languages. In scenarios where curated in-domain data is unavailable, strategies like kNN retrieval from domain-related data can approximate target data. However, this approach may not always match the level of contextual accuracy that a combination of in-domain and domain-related data offers. Although techniques like ICL and MLM improve performance, they may still fall short in capturing the intricate domain-specific nuances necessary for optimal results. As a result, models pretrained with approximate domain data may be limited by the representativeness and quality of the proxy data used.

In future work, we plan to explore whether expanding an LM’s vocabulary, coupled with increased variations of MLM, or training for a longer time, can further enhance the effectiveness of providing broad context for learning domain-specific terminology \cite{TaiKDC20,Yang2023}. Additionally, we aim to investigate the use of contrastive learning for vocabulary expansion, which could improve the model’s ability to learn domain-specific terms, especially when dealing with data that includes incomplete syntax or jargon, as commonly seen in German shift logs from the process industry \cite{Yang2023}.

\section{Conclusion}
Our research demonstrates that a new method for continual pretraining, ICL-APT, which combines in-context learning (ICL) with kNN to retrieve and augment domain-specific data using both DR and ID sources, significantly lowers the computational cost of pretraining language models while maintaining or even improving their performance. This method, tested in a resource-constrained applied setting, provides an efficient and scalable solution for domain adaptation with limited GPU resources for training and requirements for CPU usage for inference.

\section*{Acknowledgments}
This Project is supported by the Federal Ministry for Economic Affairs and Climate Action (BMWK) on the basis of a decision by the German Bundestag. Additionally, we thank Jonas L{\"u}hrs and Thomas Walton for their contributions to collecting domain-specific text data from publicly available sources.

\bibliographystyle{splncs04}
\bibliography{paper}

\begin{thebibliography}{10}
\providecommand{\url}[1]{\texttt{#1}}
\providecommand{\urlprefix}{URL }
\providecommand{\doi}[1]{https://doi.org/#1}

\bibitem{BaiRX21}
Bai, F., Ritter, A., Xu, W.: Pre-train or annotate? domain adaptation with a constrained budget. In: Proceedings of the 2021 Conference on Empirical Methods in Natural Language Processing. pp. 5002--5015. Association for Computational Linguistics, Online and Punta Cana, Dominican Republic (2021). \doi{10.18653/v1/2021.emnlp-main.409}

\bibitem{BeltagyLC19}
Beltagy, I., Lo, K., Cohan, A.: {SciBERT}: A pretrained language model for scientific text. In: Proceedings of the 2019 Conference on Empirical Methods in Natural Language Processing and the 9th International Joint Conference on Natural Language Processing (EMNLP-IJCNLP). pp. 3613--3618. Association for Computational Linguistics, Hong Kong, China (2019). \doi{10.18653/v1/D19-1371}

\bibitem{brown-2020-language}
Brown, T., Mann, B., Ryder, N., Subbiah, M., Kaplan, J.D., Dhariwal, P., Neelakantan, A., Shyam, P., Sastry, G., Askell, A., Agarwal, S., Herbert-Voss, A., Krueger, G., Henighan, T., Child, R., Ramesh, A., Ziegler, D., Wu, J., Winter, C., Hesse, C., Chen, M., Sigler, E., Litwin, M., Gray, S., Chess, B., Clark, J., Berner, C., McCandlish, S., Radford, A., Sutskever, I., Amodei, D.: Language models are few-shot learners. In: Larochelle, H., Ranzato, M., Hadsell, R., Balcan, M., Lin, H. (eds.) Advances in Neural Information Processing Systems. vol.~33, pp. 1877--1901. Curran Associates, Inc. (2020), \url{https://proceedings.neurips.cc/paper_files/paper/2020/file/1457c0d6bfcb4967418bfb8ac142f64a-Paper.pdf}

\bibitem{CaciularuCBP21}
Caciularu, A., Cohan, A., Beltagy, I., Peters, M., Cattan, A., Dagan, I.: Cdlm: Cross-document language modeling. In: Findings of the Association for Computational Linguistics: EMNLP 2021. pp. 2648--2662. Association for Computational Linguistics, Punta Cana, Dominican Republic (2021). \doi{10.18653/v1/2021.findings-emnlp.225}

\bibitem{ChanSM20}
Chan, B., Schweter, S., M{\"o}ller, T.: German's next language model. In: Proceedings of the 28th International Conference on Computational Linguistics. pp. 6788--6796. International Committee on Computational Linguistics, Barcelona, Spain (Online) (2020). \doi{10.18653/v1/2020.coling-main.598}

\bibitem{ChizhikovaLCM23}
Chizhikova, M., {L{\'o}pez-{\'U}beda}, P., {Collado-Monta{\~n}ez}, J., {Mart{\'i}n-Noguerol}, T., {D{\'i}az-Galiano}, M.C., Luna, A., {Ure{\~n}a-L{\'o}pez}, L.A., {Mart{\'i}n-Valdivia}, M.T.: {CARES}: A corpus for classification of spanish radiological reports. Computers in Biology and Medicine  \textbf{154},  106581 (Mar 2023). \doi{10.1016/j.compbiomed.2023.106581}

\bibitem{chu-wang-2018-survey}
Chu, C., Wang, R.: A survey of domain adaptation for neural machine translation. In: Bender, E.M., Derczynski, L., Isabelle, P. (eds.) Proceedings of the 27th International Conference on Computational Linguistics. pp. 1304--1319. Association for Computational Linguistics, Santa Fe, New Mexico, USA (Aug 2018), \url{https://aclanthology.org/C18-1111}

\bibitem{dnb_diss}
\firstsecond{DNB}{Deutsche National Bibliotek (DNB)}: Free online university publications. \url{https://data.dnb.de/FreieOnlineHochschulschriften/} (2024), data retrieved from the source on 2024-05-06

\bibitem{eu_patents}
\firstsecond{DPMAregister}{Deutsches Patent- und Markenamt}: {DPMAregister - Amtliche Publikations- und Registerdatenbank}. \url{https://register.dpma.de/DPMAregister/pat/basis} (2024), data retrieved from the source on 2024-01-10

\bibitem{regulations_chemistry}
\firstsecond{ECHA}{European Chemicals Agency}: Rechtsvorschriften. \url{https://echa.europa.eu/de/regulations/reach/legislation} (2024), data retrieved from the source on 2023-07-23

\bibitem{regulations_machinery}
\firstsecond{EUR-Lex}{European Parliament, Council of the European Union}: {Directive 2006/42/EC of the European Parliament and of the Council} of 17 {May} 2006 on machinery, and amending {Directive 95/16/EC}. \url{https://eur-lex.europa.eu/legal-content/EN/ALL/?uri=celex\%3A32006L0042} (2019), data retrieved from the source on 2023-07-26

\bibitem{regulations_pharma}
\firstsecond{European Comission}{European Comission. Directorate-General for Health and Food Safety}: Legal framework governing medicinal products for human use in the {EU}. \url{https://health.ec.europa.eu/medicinal-products/legal-framework-governing-medicinal-products-human-use-eu_en?etrans=de&prefLang=de} (2024), data retrieved from the source on 2023-07-24

\bibitem{ddr_patents}
\firstsecond{GESIS}{GESIS – Leibniz-Institut für Sozialwissenschaften}: {Patentdaten der Deutschen Demokratischen Republik (DDR) }(1949-1990). \url{https://doi.org/10.7802/2423} (2022), data retrieved from the source on 2023-08-12

\bibitem{GargMS24}
Garg, S., Moghaddam, R.Z., Sundaresan, N.: {RAPGen}: An approach for fixing code inefficiencies in zero-shot (Jul 2024), \url{https://arxiv.org/abs/2306.17077}

\bibitem{GunasekarZAM23}
Gunasekar, S., Zhang, Y., Aneja, J., Mendes, C.C.T., Del~Giorno, A., Gopi, S., Javaheripi, M., Kauffmann, P., {de Rosa}, G., Saarikivi, O., Salim, A., Shah, S., Behl, H.S., Wang, X., Bubeck, S., Eldan, R., Kalai, A.T., Lee, Y.T., Li, Y.: Textbooks are all you need (Oct 2023)

\bibitem{GururanganMSL20}
Gururangan, S., Marasovi{\'c}, A., Swayamdipta, S., Lo, K., Beltagy, I., Downey, D., Smith, N.A.: Don't stop pretraining: Adapt language models to domains and tasks. In: Proceedings of the 58th Annual Meeting of the Association for Computational Linguistics. pp. 8342--8360. Association for Computational Linguistics, Online (2020). \doi{10.18653/v1/2020.acl-main.740}

\bibitem{hedderich-etal-2021-survey}
Hedderich, M.A., Lange, L., Adel, H., Str{\"o}tgen, J., Klakow, D.: A survey on recent approaches for natural language processing in low-resource scenarios. In: Toutanova, K., Rumshisky, A., Zettlemoyer, L., Hakkani-Tur, D., Beltagy, I., Bethard, S., Cotterell, R., Chakraborty, T., Zhou, Y. (eds.) Proceedings of the 2021 Conference of the North American Chapter of the Association for Computational Linguistics: Human Language Technologies. pp. 2545--2568. Association for Computational Linguistics, Online (Jun 2021). \doi{10.18653/v1/2021.naacl-main.201}, \url{https://aclanthology.org/2021.naacl-main.201}

\bibitem{JiangJXZ23}
Jiang, G., Jiang, C., Xue, S., Zhang, J., Zhou, J., Lian, D., Wei, Y.: Towards anytime fine-tuning: Continually pre-trained language models with hypernetwork prompts. In: Findings of the Association for Computational Linguistics: EMNLP 2023. pp. 12081--12095. Association for Computational Linguistics, Singapore (2023). \doi{10.18653/v1/2023.findings-emnlp.808}

\bibitem{joshi-etal-2020-state}
Joshi, P., Santy, S., Budhiraja, A., Bali, K., Choudhury, M.: The state and fate of linguistic diversity and inclusion in the {NLP} world. In: Jurafsky, D., Chai, J., Schluter, N., Tetreault, J. (eds.) Proceedings of the 58th Annual Meeting of the Association for Computational Linguistics. pp. 6282--6293. Association for Computational Linguistics, Online (Jul 2020). \doi{10.18653/v1/2020.acl-main.560}, \url{https://aclanthology.org/2020.acl-main.560/}

\bibitem{KeSLK23}
Ke, Z., Shao, Y., Lin, H., Konishi, T., Kim, G., Liu, B.: Continual pre-training of language models. In: The Eleventh International Conference on Learning Representations, ICLR 2023, Kigali, Rwanda, May 1-5, 2023. OpenReview.net (2023), \url{https://openreview.net/forum?id=m\_GDIItaI3o}

\bibitem{LongWP23}
Long, Q., Wang, W., Pan, S.: Adapt in contexts: Retrieval-augmented domain adaptation via in-context learning. In: Proceedings of the 2023 Conference on Empirical Methods in Natural Language Processing. pp. 6525--6542. Association for Computational Linguistics, Singapore (2023). \doi{10.18653/v1/2023.emnlp-main.402}

\bibitem{MonajatipoorYSE24}
Monajatipoor, M., Yang, J., Stremmel, J., Emami, M., Mohaghegh, F., Rouhsedaghat, M., Chang, K.W.: Llms in biomedicine: A study on clinical named entity recognition (Apr 2024)

\bibitem{mosbach-etal-2023-shot}
Mosbach, M., Pimentel, T., Ravfogel, S., Klakow, D., Elazar, Y.: Few-shot fine-tuning vs. in-context learning: A fair comparison and evaluation. In: Rogers, A., Boyd-Graber, J., Okazaki, N. (eds.) Findings of the Association for Computational Linguistics: ACL 2023. pp. 12284--12314. Association for Computational Linguistics, Toronto, Canada (Jul 2023). \doi{10.18653/v1/2023.findings-acl.779}, \url{https://aclanthology.org/2023.findings-acl.779/}

\bibitem{Tri2016}
Nguyen, T., Rosenberg, M., Song, X., Gao, J., Tiwary, S., Majumder, R., Deng, L.: {MS} {MARCO:} {A} human generated machine reading comprehension dataset. In: Besold, T.R., Bordes, A., d'Avila Garcez, A.S., Wayne, G. (eds.) Proceedings of the Workshop on Cognitive Computation: Integrating neural and symbolic approaches 2016 co-located with the 30th Annual Conference on Neural Information Processing Systems {(NIPS} 2016), Barcelona, Spain, December 9, 2016. {CEUR} Workshop Proceedings, vol.~1773. CEUR-WS.org (2016), \url{https://ceur-ws.org/Vol-1773/CoCoNIPS\_2016\_paper9.pdf}

\bibitem{ReimersG20}
Reimers, N., Gurevych, I.: Making monolingual sentence embeddings multilingual using knowledge distillation. In: Proceedings of the 2020 Conference on Empirical Methods in Natural Language Processing (EMNLP). pp. 4512--4525. Association for Computational Linguistics, Online (2020). \doi{10.18653/v1/2020.emnlp-main.365}

\bibitem{RojasDV22}
Rojas, M., Dunstan, J., Villena, F.: Clinical {Flair}: A pre-trained language model for spanish clinical natural language processing. In: Proceedings of the 4th Clinical Natural Language Processing Workshop. pp. 87--92. Association for Computational Linguistics, Seattle, WA (2022). \doi{10.18653/v1/2022.clinicalnlp-1.9}

\bibitem{TaiKDC20}
Tai, W., Kung, H.T., Dong, X., Comiter, M., Kuo, C.F.: {exBERT}: Extending pre-trained models with domain-specific vocabulary under constrained training resources. In: Findings of the Association for Computational Linguistics: EMNLP 2020. pp. 1433--1439. Association for Computational Linguistics, Online (2020). \doi{10.18653/v1/2020.findings-emnlp.129}

\bibitem{beir_2021}
Thakur, N., Reimers, N., R\"{u}ckl\'{e}, A., Srivastava, A., Gurevych, I.: Beir: A heterogeneous benchmark for zero-shot evaluation of information retrieval models. In: Vanschoren, J., Yeung, S. (eds.) Proceedings of the Neural Information Processing Systems Track on Datasets and Benchmarks. vol.~1 (2021), \url{https://datasets-benchmarks-proceedings.neurips.cc/paper_files/paper/2021/file/65b9eea6e1cc6bb9f0cd2a47751a186f-Paper-round2.pdf}

\bibitem{XieSML23}
Xie, S.M., Santurkar, S., Ma, T., Liang, P.: Data selection for language models via importance resampling. In: Proceedings of the 37th International Conference on Neural Information Processing Systems. NIPS '23, Curran Associates Inc., Red Hook, NY, USA (2023)

\bibitem{Yang2023}
Yang, J., Hu, X., Huang, W., Yuan, H., Shen, Y., Xiao, G.: Advancing domain adaptation of {BERT} by learning domain term semantics. In: Knowledge Science, Engineering and Management: 16th International Conference, KSEM 2023, Guangzhou, China, August 16–18, 2023, Proceedings, Part IV. p. 12–24. Springer-Verlag, Berlin, Heidelberg (2023). \doi{10.1007/978-3-031-40292-0_2}, \url{https://doi.org/10.1007/978-3-031-40292-0_2}

\bibitem{zhukova-etal-2025-automated}
Zhukova, A., Matt, C.E., Gipp, B.: Automated collection of evaluation dataset for semantic search in low-resource domain language. In: Hettiarachchi, H., Ranasinghe, T., Rayson, P., Mitkov, R., Gaber, M., Premasiri, D., Tan, F.A., Uyangodage, L. (eds.) Proceedings of the First Workshop on Language Models for Low-Resource Languages. pp. 112--122. Association for Computational Linguistics, Abu Dhabi, United Arab Emirates (Jan 2025), \url{https://aclanthology.org/2025.loreslm-1.8/}

\bibitem{Zhukova2024}
Zhukova, A., von Sperl, L., Matt, C.E., Gipp, B.: Generative user-experience research for developing domain-specific natural language processing applications. Knowledge and Information Systems  \textbf{66},  7859–7889 (September 2024). \doi{10.1007/s10115-024-02212-5}, \url{https://doi.org/10.1007/s10115-024-02212-5}

\end{thebibliography}

\end{document}